\documentclass{article}
\usepackage[table]{xcolor}
\usepackage[preprint]{neurips_2026}

\usepackage[utf8]{inputenc} 
\usepackage[T1]{fontenc}    
\usepackage{hyperref}       
\usepackage{url}            
\usepackage{booktabs}       
\usepackage{amsfonts}       
\usepackage{nicefrac}       
\usepackage{microtype}      
\usepackage{xcolor}         

\usepackage{graphicx}
\usepackage{amssymb}
\usepackage{amsmath}
\usepackage{algorithm}
\usepackage{algpseudocode}
\usepackage{booktabs}
\usepackage{colortbl}
\usepackage{placeins}
\usepackage{dblfloatfix}
\usepackage[most]{tcolorbox}
\usepackage{enumitem}
\definecolor{bopsdrowblue}{RGB}{232,244,251}
\definecolor{briefboxbg}{RGB}{247,250,252}
\definecolor{briefboxline}{RGB}{111,149,171}
\definecolor{briefboxtitle}{RGB}{226,238,244}
\newtcolorbox{briefexample}[1]{enhanced,title={#1},
  colback=briefboxbg,colframe=briefboxline,colbacktitle=briefboxtitle,
  coltitle=black,fonttitle=\bfseries,fontupper=\small,boxrule=0.7pt,arc=2mm,
  left=6pt,right=6pt,top=5pt,bottom=5pt,
  before skip=8pt,after skip=8pt,
  before upper={\setlength{\parindent}{0pt}\setlength{\parskip}{2pt}}}
\definecolor{tableyes}{RGB}{38,111,101}
\definecolor{tableno}{RGB}{121,132,140}
\newcommand{\tableyes}{\textcolor{tableyes}{$\checkmark$}}
\newcommand{\tableno}{\textcolor{tableno}{$\times$}}

\title{B-OPSD: On-Policy Self-Distillation with Self-Generated Problem Briefs}

\author{%
  \textbf{Zhi Xu}\thanks{Equal contribution.}\quad
  \textbf{Yinqi Wu}\footnotemark[1]\quad
  \textbf{Feng Zhang} \quad
  \textbf{Xiaotong Zhang}\quad
  \textbf{Hong Yu}\quad
  \textbf{Han Liu}\thanks{Corresponding author.}\\
  Dalian University of Technology\\
  Dalian, China\\
  \texttt{xu.zhi.dut@gmail.com, wuyinqi@mail.dlut.edu.cn}\\
  \texttt{fengzhangyvonne@gmail.com, zxt.dut@hotmail.com}\\
  \texttt{hongyu@dlut.edu.cn, liu.han.dut@gmail.com}%
}

\begin{document}

\maketitle

\begin{abstract}
On-Policy Self-Distillation (OPSD) has recently attracted interest because it provides dense teacher guidance without a separate, often larger teacher model. However, OPSD conditions its teacher on ground-truth solutions, which limits its use when such solutions are unavailable. We propose On-Policy Self-Distillation with Self-Generated Problem Briefs (B-OPSD). Using a prompt template, the initial model or a saved student checkpoint generates a problem brief from the question alone. The prompt asks for relevant principles, applicability conditions, and likely pitfalls, but not a final answer or a complete solution. A frozen teacher uses the brief as privileged context to guide the current student's on-policy rollouts through token-level distribution matching. B-OPSD uses no ground-truth solutions during training. Results on AIME 2024, AIME 2025, and HMMT 2025 show that both B-OPSD variants achieve higher average accuracy than OPSD at each of three Qwen3 model sizes. Code is available at \url{https://github.com/ZevineXu/B-OPSD}.
\end{abstract}

\section{Introduction}
Large language models have shown strong performance in language understanding \citep{cao2026tablemaster,Lee2026ERGO,zhang2026towards}, mathematical reasoning \citep{wei2022chain,lewkowycz2022solving,chen2025seedprover}, and code generation \citep{openai2023gpt4,Kang2025Distilling,Dai2025Meshcoder}. Their performance on specific tasks can be improved through supervised fine-tuning (SFT), reinforcement learning with verifiable rewards (RLVR), and on-policy distillation (OPD) \citep{ouyang2022training,shao2024deepseekmath,agarwal2024onpolicy}.

These methods provide different learning signals. SFT trains the model on reference responses supplied in the training data. In mathematical reasoning, RLVR can sample responses from the current model and score their final answers, but an outcome reward does not identify which reasoning steps caused an error. OPD also uses responses sampled from the student. A teacher then evaluates each student prefix and provides guidance through its next-token distribution \citep{agarwal2024onpolicy,lu2025onpolicy}.

OPD commonly uses a separate teacher model to provide this guidance. On-Policy Self-Distillation (OPSD) instead gives the same base model two roles with different input contexts \citep{zhao2026opsd}. Its student sees the problem, while its teacher also sees a reference solution. Both roles evaluate the student's trajectory, and the teacher's distribution guides the student at each token. OPSD does not use a separate teacher model during distillation, but it still needs a reference solution for each training problem. Preparing reference solutions for a dataset of questions alone may require an external model and validation. This dependence raises a practical question: \textbf{Can on-policy self-distillation work when no correct reference solution is available?}

\begin{figure*}[t]
  \centering
  \includegraphics[width=0.95\textwidth]{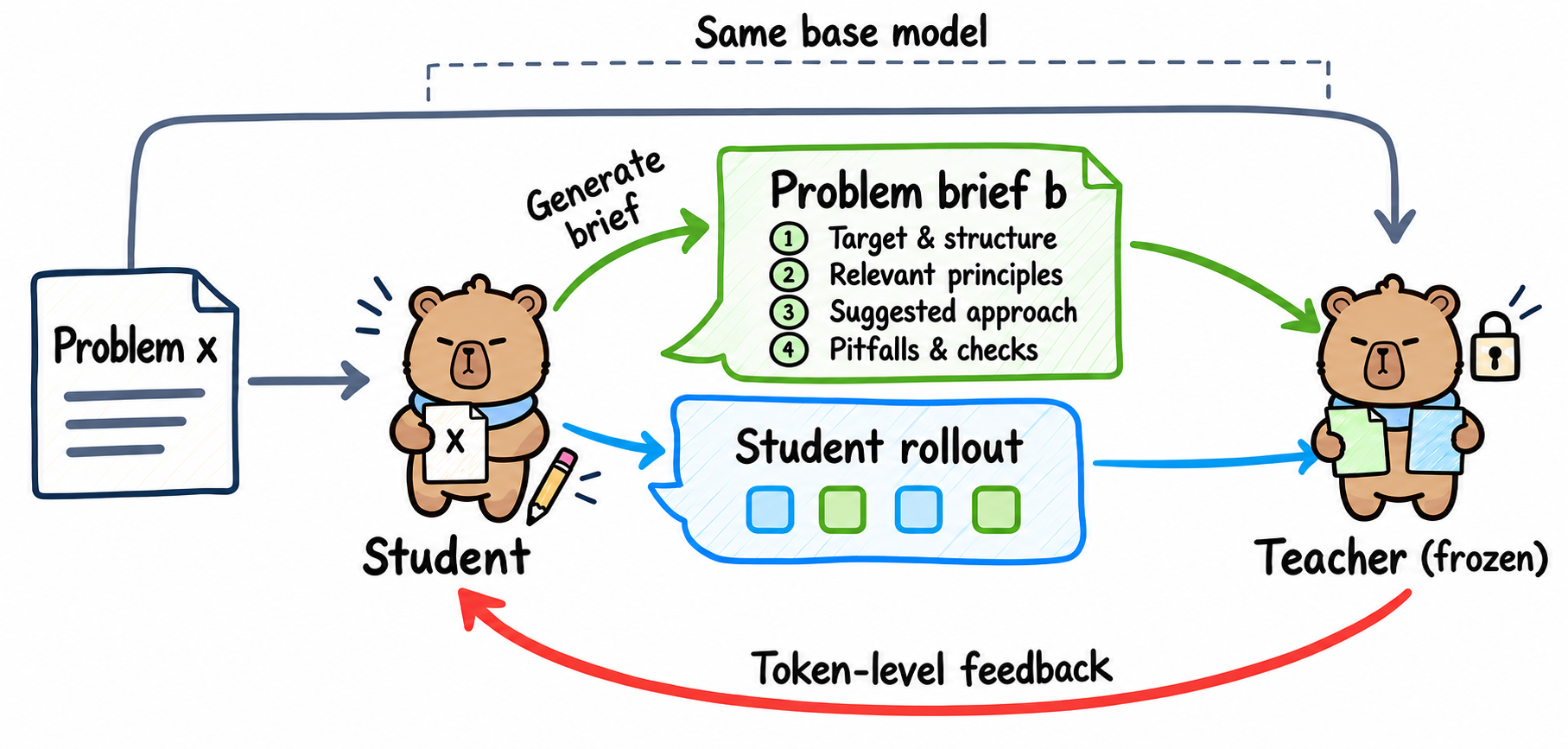}
  \caption{Overview of B-OPSD. Student and teacher start from the same base model; the teacher stays frozen while the student is updated. The initial model generates static briefs, while a saved student checkpoint generates refreshed briefs. The current student generates its solution trajectory from the question alone. The frozen teacher receives the question and brief, evaluates the same student prefixes, and provides token-level feedback.}
  \label{fig:bopsd-overview}
\end{figure*}

To answer this question, we propose On-Policy Self-Distillation with Self-Generated Problem Briefs (B-OPSD). The model generates a \emph{problem brief} from the question alone. The brief is intended to connect relevant principles to the problem, state when they apply, and note likely pitfalls and checks. The prompt instructs the model not to provide a final answer or a complete solution. As Figure~\ref{fig:bopsd-overview} shows, the student samples a solution trajectory from the question, while the teacher evaluates the same trajectory with the brief as additional context. Their token-level distribution difference updates only the student. We consider briefs generated once before training and briefs refreshed from earlier student checkpoints. Neither version uses a reference solution or an external model to prepare teacher context. On AIME 2024, AIME 2025, and HMMT 2025, both versions achieve higher average accuracy across the three benchmarks than OPSD at each of three Qwen3 model sizes in the reported runs. These results suggest that self-generated problem briefs can provide useful teacher context without a verified solution.

\section{Related Work}

\paragraph{Supervised fine-tuning (SFT).}
SFT trains on reference responses, predicting each token from the preceding reference tokens \citep{ouyang2022training,Yao2025Hier,fan2025preference}. Sequence-level KD uses teacher responses as targets \citep{kim2016sequence}, while STaR and ReST$^{\mathrm{EM}}$ train on solutions selected with answer feedback \citep{zelikman2022star,singh2024beyond}. Training on reference prefixes but generating from the model's own prefixes at inference creates exposure bias.

\paragraph{Reinforcement learning with verifiable rewards (RLVR).}
RLVR samples responses from the current policy and scores their final answers.\cite{schulman2017proximal,xi2025bapo,zhang2026entropy} PPO uses clipped updates but typically trains a value model, which adds memory cost \citep{schulman2017proximal}. GRPO avoids this model by estimating advantages from groups of responses \citep{shao2024deepseekmath}. Building on GRPO, DAPO raises the upper clipping bound and filters groups with uniform rewards \citep{yu2025dapo}. GSPO replaces GRPO's token-level importance ratios and clipping with sequence-level versions \citep{zheng2025gspo}. Yet outcome rewards cannot locate a reasoning error or provide a teacher distribution at each token.

\paragraph{On-policy distillation (OPD).}
OPD addresses the limitation of RLVR by matching a teacher's next-token distributions on trajectories generated by the current student \citep{lu2025onpolicy,yuan2026visionopd,Xueruwen2025OPS}. GKD trains on student-generated sequences and allows different divergences to match the teacher and student distributions \citep{agarwal2024onpolicy}. Entropy-Aware OPD adds forward KL to the reverse KL loss when the teacher distribution has high entropy \citep{jin2026entropy}. Many existing OPD systems use a separate, stronger teacher model. The teacher is often larger than the student, which increases training cost.

\begin{table*}[ht]
  \centering
  \caption{B-OPSD combines on-policy student rollouts and token-level teacher feedback without a separate teacher model, reference solutions, or verified answers during training. The SFT and GRPO rows use the reference solutions and verified final answers, respectively, from the compared training setups. The OPD row represents a common setup with a separate teacher model and no answer labels during distillation.}
  \begin{tabular}{lcccc}
    \toprule
    & \multicolumn{2}{c}{\textit{Training signal}} & \multicolumn{2}{c}{\textit{Training requirements}} \\
    \cmidrule(lr){2-3}\cmidrule(l){4-5}
    Method & \shortstack{On-policy\\data} & \shortstack{Dense learning\\signal} & \shortstack{No separate\\teacher model} & \shortstack{No answer\\labels} \\
    \midrule
    SFT & \tableno & \tableyes & \tableyes & \tableno \\
    GRPO & \tableyes & \tableno & \tableyes & \tableno \\
    OPD & \tableyes & \tableyes & \tableno & \tableyes \\
    OPSD & \tableyes & \tableyes & \tableyes & \tableno \\
    \midrule
    \rowcolor{bopsdrowblue} \textbf{B-OPSD} & \tableyes & \tableyes & \tableyes & \tableyes \\
    \bottomrule
  \end{tabular}
  \label{tab:method-comparison}
\end{table*}

\paragraph{On-policy self-distillation (OPSD).}
OPSD gives student and teacher roles to the same base model, with a reference answer or solution available only to the teacher \citep{zhao2026opsd}. Context distillation also uses extra information during training, and On-Policy Context Distillation conditions a teacher on historical solutions or optimized prompts \citep{snell2022learning,ye2026opcd}. OPSD removes the separate teacher model but still needs a reference for each training problem. B-OPSD replaces that reference with a self-generated problem brief. The teacher uses the brief to guide student rollouts through token distributions; no reference answer is read during training. Table~\ref{tab:method-comparison} compares the training setups.

\section{Method}
\label{sec:method}

\subsection{Generating Problem Briefs}
\label{sec:briefs}

\begin{figure*}[t]
  \centering
  \includegraphics[width=0.9\textwidth]{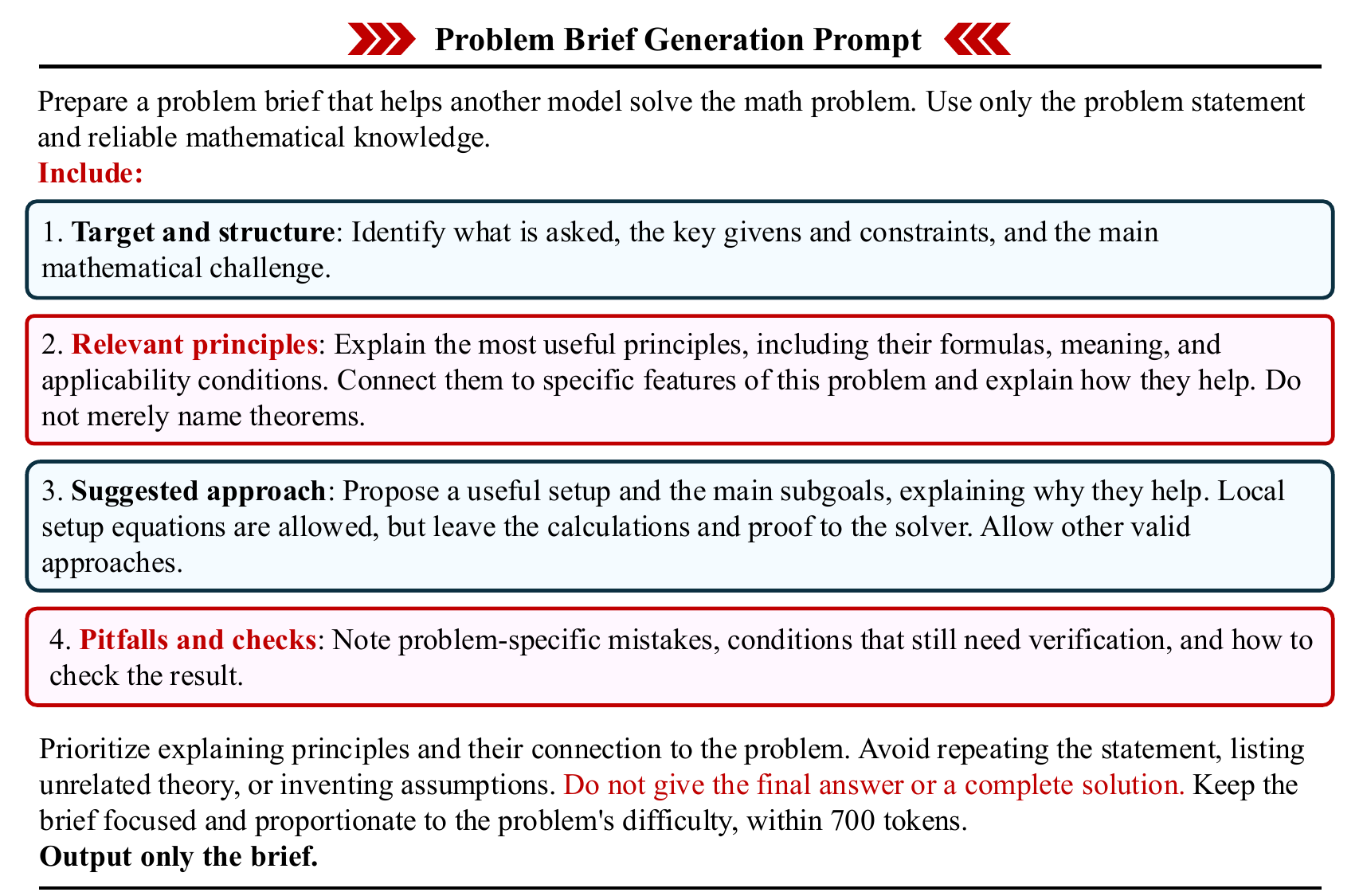}
  \caption{Complete problem brief generation prompt $\mathcal P_B$. The same prompt is used for static and refreshed briefs. The generator also receives the problem statement $x$.}
  \label{fig:brief-prompt}
\end{figure*}

We train B-OPSD on a set $\mathcal D=\{x_i\}_{i=1}^{N}$ of mathematical problems without reading reference solutions or using an outcome verifier. OPSD supplies its teacher with a reference solution \citep{zhao2026opsd}. B-OPSD instead prepares teacher context from the problem itself. Identifying relevant principles and checking their conditions may be possible even when producing a complete correct solution is difficult. We use a problem brief to make this information available to the teacher. Figure~\ref{fig:brief-prompt} shows the complete generation prompt $\mathcal P_B$.

The prompt has four parts. First, it asks the generator to identify the target, givens, constraints, and main challenge. Second, it asks for relevant principles, including their formulas, meaning, applicability conditions, and connection to the problem. Third, it asks for a useful setup and main subgoals. Local setup equations are allowed, but the calculations and proof are left to the solver, and other valid approaches remain possible. Fourth, it asks for problem-specific pitfalls, conditions to verify, and ways to check the result. The prompt also asks the generator to avoid unrelated theory and invented assumptions, omit the final answer and complete solution, keep the brief within 700 tokens, and output only the brief.

Let $\theta_0$ be the initial model parameters and $\theta_k$ the student parameters after $k$ updates. The brief generator has parameters $\phi_k$ from the initial model or a saved student checkpoint. Given a problem $x$, it samples a brief as
\begin{equation}
    b \sim p_{\phi_k}(\cdot\mid\mathcal P_B,x).
    \label{eq:brief-generation}
\end{equation}
We use one brief for each problem when it is processed and pass the brief to the teacher as text. Static briefs are reused, while refreshed briefs are generated using the selected checkpoint. Gradients do not pass through brief generation. We do not verify or filter generated briefs. Appendix~\ref{sec:brief-examples} shows outputs from the generation records.

The teacher prompt presents the original problem followed by the brief and asks the model to keep each reasoning step valid. The student receives the problem without the brief. The brief is teacher context, not a target response for the student to reproduce. It is not checked against a reference solution and may contain mistakes. The method uses it to change the teacher's conditioning context, not as a verified answer.

\subsection{On-Policy Self-Distillation with Problem Briefs}
\label{sec:brief-distillation}

We retain a frozen teacher with parameters $\bar\theta=\theta_0$ and update only the student. At training step $k$, the student samples a solution trajectory from its current policy:
\begin{equation}
    \hat y\sim p_{\theta_k}(\cdot\mid x).
    \label{eq:student-rollout}
\end{equation}
Let $T=|\hat y|$ be its length and $\hat y_{<t}$ its prefix at position $t$. The student and teacher evaluate the same prefix, with the brief available only to the teacher:
\begin{align}
    p_{S,t}^{\theta} &= p_{\theta}(\cdot\mid x,\hat y_{<t}),
    \label{eq:student-distribution}\\
    p_{T,t} &= p_{\bar\theta}(\cdot\mid x,b,\hat y_{<t}).
    \label{eq:teacher-distribution}
\end{align}
The role-specific prompts are omitted for clarity. To evaluate a trajectory, we append the same sampled prefix to each role's prompt and compute next-token distributions. The teacher does not need to sample a separate solution. Its forward pass supplies a distribution at every position of the student's trajectory. Thus, the loss can be computed for a trajectory without first judging its final answer. The teacher's guidance still depends on the quality of the generated brief; it does not certify that a step or answer is correct.

\begin{algorithm}[t]
\caption{B-OPSD training with static or refreshed problem briefs}
\label{alg:bopsd}
\begin{algorithmic}[1]
\Require Problems $\mathcal D$, initial parameters $\theta_0$, brief prompt $\mathcal P_B$, training steps $K$
\Require Brief mode, checkpoint schedule $\mathcal C$, divergence $D$, threshold $\tau$
\State Freeze teacher at $\bar\theta\gets\theta_0$
\If{brief mode is static}
    \ForAll{$x\in\mathcal D$}
        \State Generate and store $b_0(x)\sim p_{\theta_0}(\cdot\mid\mathcal P_B,x)$
    \EndFor
\EndIf
\For{$k=0,1,\ldots,K-1$}
    \State Sample a batch $\mathcal B_k$ from $\mathcal D$
    \If{brief mode is refreshed}
        \State Select generator $\phi_k\gets\theta_{s(k)}$ using Eq.~\ref{eq:brief-checkpoint}
    \EndIf
    \ForAll{$x\in\mathcal B_k$}
        \If{brief mode is static}
            \State $b_k(x)\gets b_0(x)$
        \Else
            \State $b_k(x)\sim p_{\phi_k}(\cdot\mid\mathcal P_B,x)$
        \EndIf
        \State Sample a student trajectory $\hat y_x\sim p_{\theta_k}(\cdot\mid x)$
        \State Evaluate student and teacher distributions on $\hat y_x$ using Eqs.~\ref{eq:student-distribution}--\ref{eq:teacher-distribution}
    \EndFor
    \State $\mathcal L_k(\theta)\gets\dfrac{1}{|\mathcal B_k|}\sum_{x\in\mathcal B_k}\ell_D^\tau(\theta;x,b_k(x),\hat y_x)$
    \State Update only the student to obtain $\theta_{k+1}$ by minimizing $\mathcal L_k$
    \If{$k+1\in\mathcal C$}
        \State Save student checkpoint $\theta_{k+1}$
    \EndIf
\EndFor
\State \Return $p_{\theta_K}$
\end{algorithmic}
\end{algorithm}

\paragraph{Distribution matching.}
For a fixed brief and sampled trajectory, we average the divergence between the two next-token distributions over the positions in the trajectory:
\begin{equation}
    \ell_D(\theta;x,b,\hat y)
    =\frac{1}{T}\sum_{t=1}^{T}
    D\!\left(p_{T,t}\,\|\,p_{S,t}^{\theta}\right).
    \label{eq:brief-loss}
\end{equation}
The loss compares full distributions rather than only the probabilities of the sampled tokens. We consider three choices for $D$ in Eq.~\ref{eq:brief-loss}. Forward KL weights the log ratio by the teacher distribution, while reverse KL weights it by the student distribution:
\begin{align}
    D_{\mathrm{FKL}}
    &=D_{\mathrm{KL}}(p_{T,t}\|p_{S,t}^{\theta}),
    \label{eq:forward-kl}\\
    D_{\mathrm{RKL}}
    &=D_{\mathrm{KL}}(p_{S,t}^{\theta}\|p_{T,t}),
    \label{eq:reverse-kl}
\end{align}
JSD compares each distribution with their mixture $m_t=\beta p_{T,t}+(1-\beta)p_{S,t}^{\theta}$:
\begin{align}
    D_{\mathrm{JSD}_{\beta}}
    &=\beta D_{\mathrm{KL}}(p_{T,t}\|m_t) +(1-\beta)D_{\mathrm{KL}}(p_{S,t}^{\theta}\|m_t),
    \label{eq:brief-jsd}
\end{align}
Here $D_{\mathrm{KL}}$ denotes KL divergence and $\beta\in[0,1]$. We use $\beta=0.5$ in the divergence comparison. The main experiments use forward KL. In the divergence comparison, we train a separate model with each loss under the same teacher conditioning.

\paragraph{Pointwise divergence clipping.}
We retain the pointwise clipping procedure of OPSD \citep{zhao2026opsd}. Let $\mathcal V$ be the model vocabulary and $d_{t,v}^{D}$ the contribution of entry $v$ to the selected divergence at position $t$, so that
\begin{equation}
    D\!\left(p_{T,t}\,\|\,p_{S,t}^{\theta}\right)
    =\sum_{v\in\mathcal V}d_{t,v}^{D}.
\end{equation}
For a threshold $\tau$, the clipped loss is
\begin{equation}
    \ell_D^{\tau}(\theta;x,b,\hat y)
    =\frac{1}{T}\sum_{t=1}^{T}\sum_{v\in\mathcal V}
      \min\!\left(d_{t,v}^{D},\tau\right).
    \label{eq:clipped-brief-loss}
\end{equation}
For forward KL, $d_{t,v}^{D}$ is the teacher probability $p_{T,t}(v)$ times the log ratio of teacher to student probabilities at $v$. Reverse KL exchanges the teacher and student probabilities. For JSD, $d_{t,v}^{D}$ is the sum of the two weighted KL contributions in Eq.~\ref{eq:brief-jsd}, evaluated at $v$. Clipping is applied to these vocabulary contributions before they are summed over entries and positions. It does not clip the completed sequence loss. The divergence choice and clipping threshold are reported in the experimental settings.

For a batch $\mathcal B_k$, we average the clipped losses:
\begin{equation}
    \mathcal L_k(\theta)
    =\frac{1}{|\mathcal B_k|}
      \sum_{x\in\mathcal B_k}
      \ell_D^{\tau}\!\left(\theta;x,b_k(x),\hat y_x\right),
    \label{eq:batch-loss}
\end{equation}
where $b_k(x)$ is the brief assigned to $x$ at step $k$ and $\hat y_x$ is sampled according to Eq.~\ref{eq:student-rollout}. The optimizer updates $\theta_k$ using this loss. Neither the teacher nor the brief generator receives gradients from the update.
The sampled trajectory and teacher probabilities are held fixed during this update; gradients pass only through the student probabilities. A new trajectory is sampled from the updated student at later steps. Thus, the rollouts remain on-policy while the teacher stays at $\theta_0$.

\subsection{Static and Refreshed Briefs}
\label{sec:brief-variants}

We consider two ways to choose the brief generator. Both use the same generation prompt, frozen teacher, and distillation procedure. They differ in whether the generator remains at the initial model or follows saved student checkpoints.

\paragraph{B-OPSD with static briefs.}
Before training, the initial model generates one brief for every training problem. We retain each generated brief without format filtering and reuse it throughout training. Thus, $\phi_k=\theta_0$ for all $k$, and $b_k(x)=b_0(x)$. This variant keeps the teacher context fixed for each problem. Its teacher distributions can still change across visits to that problem because the student may generate different trajectory prefixes.

\paragraph{B-OPSD with refreshed briefs.}
The generator uses the most recently saved student checkpoint. Let $\mathcal C$ be the set of completed update counts at which checkpoints are saved, including $0$. At step $k$, we select
\begin{equation}
    s(k)=\max\{s\in\mathcal C:s\leq k\},
    \qquad \phi_k=\theta_{s(k)}.
    \label{eq:brief-checkpoint}
\end{equation}
The initial model supplies briefs until a later checkpoint becomes available. Subsequent briefs use the selected checkpoint. This allows the brief content to reflect changes in the student during training without updating the teacher parameters. The checkpoint interval and generation settings are specified in the experiments. 

Algorithm~\ref{alg:bopsd} summarizes the training procedure. In both variants, the current student generates the trajectories used for distillation. The source of the brief does not change this on-policy property. At inference, we use the trained student with the problem alone; brief generation and teacher evaluation are training operations.

\section{Experiments}

\subsection{Experimental Setup}

\begin{table*}[t]
  \centering
  \caption{Avg@12 accuracy (\%) across three benchmarks. B-OPSD uses static briefs; B-OPSD$^\star$ uses refreshed briefs. Average is computed from unrounded scores. Shaded rows show the B-OPSD variants, and bold marks the highest score for each model size and column.}
  \begin{tabular}{lcccc}
    \toprule
    \textbf{Method} & \textbf{AIME 2024} & \textbf{AIME 2025} & \textbf{HMMT 2025} & \textbf{Average} \\
    \midrule
    \multicolumn{5}{l}{\textbf{Qwen3-1.7B}} \\
    Base & 51.5 & 36.7 & 23.1 & 37.1 \\
    SFT & 48.4 & 36.3 & 22.7 & 35.8 \\
    GRPO & 51.1 & 38.3 & 23.7 & 37.7 \\
    OPSD & 54.7 & \textbf{43.1} & 25.0 & 40.9 \\
    \rowcolor{bopsdrowblue} B-OPSD & 56.4 & 42.2 & 27.2 & 41.9 \\
    \rowcolor{bopsdrowblue} B-OPSD$^\star$ & \textbf{57.5} & 42.5 & \textbf{28.6} & \textbf{42.9} \\
    \midrule
    \multicolumn{5}{l}{\textbf{Qwen3-4B}} \\
    Base & 74.9 & 66.4 & 42.2 & 61.2 \\
    SFT & 70.2 & 62.3 & 43.4 & 58.6 \\
    GRPO & 75.6 & 68.1 & 44.4 & 62.7 \\
    OPSD & 75.6 & 67.5 & 45.8 & 63.0 \\
    \rowcolor{bopsdrowblue} B-OPSD & 76.4 & 68.3 & \textbf{46.1} & 63.6 \\
    \rowcolor{bopsdrowblue} B-OPSD$^\star$ & \textbf{77.5} & \textbf{70.3} & 45.0 & \textbf{64.3} \\
    \midrule
    \multicolumn{5}{l}{\textbf{Qwen3-8B}} \\
    Base & 75.8 & 65.6 & 43.9 & 61.8 \\
    SFT & 72.3 & 64.2 & 42.9 & 59.8 \\
    GRPO & 76.4 & 68.9 & 46.7 & 64.0 \\
    OPSD & 77.8 & \textbf{71.9} & 46.7 & 65.5 \\
    \rowcolor{bopsdrowblue} B-OPSD & \textbf{79.7} & 71.1 & 47.5 & 66.1 \\
    \rowcolor{bopsdrowblue} B-OPSD$^\star$ & 78.3 & \textbf{71.9} & \textbf{50.0} & \textbf{66.8} \\
    \bottomrule
  \end{tabular}
  \label{tab:main-results}
\end{table*}

\begin{figure*}[t]
  \centering
  \includegraphics[width=0.98\textwidth]{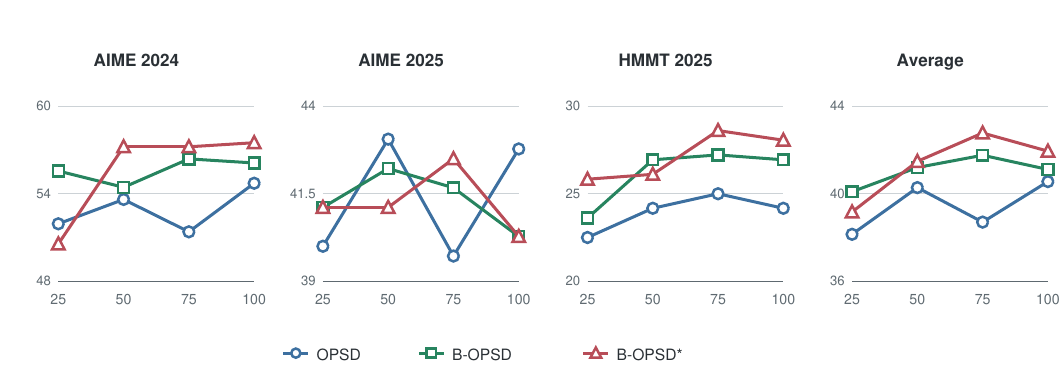}
  \caption{Checkpoint Avg@12 accuracy (\%) for Qwen3-1.7B. B-OPSD uses static briefs; B-OPSD$^\star$ uses refreshed briefs. The Average panel combines scores from the same checkpoint; Table~\ref{tab:main-results} averages each benchmark's best score. The vertical axes start near the observed scores and differ across panels.}
  \label{fig:checkpoint-1p7b}
\end{figure*}

\paragraph{Models and data.}
Following OPSD \citep{zhao2026opsd}, we use the instruct-tuned Qwen3-1.7B, Qwen3-4B, and Qwen3-8B models \citep{yang2025qwen3}. We sample up to 30,000 mathematical problems from the OpenThoughts reasoning subset \citep{guha2025openthoughts}. B-OPSD reads only the problem statements during training. We evaluate on AIME 2024, AIME 2025, and HMMT 2025.

\paragraph{Baselines.}
We compare B-OPSD with the initial model, SFT, GRPO, and OPSD. We take the Base, SFT, and GRPO results from OPSD~\cite{zhao2026opsd} and reproduce OPSD ourselves. In the original study, SFT trains on the reference solutions in OpenThoughts, while GRPO uses verified final answers to provide outcome rewards. Our OPSD reproduction uses the reference solutions as privileged teacher context.

\paragraph{Implementation details.}
Our OPSD reproduction and B-OPSD runs use LoRA on NVIDIA A800 GPUs \citep{hu2022lora}. Following OPSD, we use a learning rate of $5\times10^{-6}$, an effective batch size of 32, LoRA rank 64, and LoRA scaling factor 128. All three distillation runs train for 100 updates. We sample one student trajectory per problem at temperature $1.1$, with a maximum length of 1,024 tokens. The teacher remains fixed at the initial model while the student is updated. The main B-OPSD runs use full-vocabulary forward KL with pointwise clipping at $\tau=0.05$.

We generate problem briefs with temperature $0.0$. B-OPSD with static briefs generates each brief before training. B-OPSD with refreshed briefs uses the most recently saved student checkpoint and saves a new checkpoint every 25 updates. Appendix~\ref{sec:additional-implementation} gives further implementation details.

\paragraph{Evaluation.}
Following OPSD, our evaluations report Avg@12 accuracy from 12 samples per problem. We use thinking mode, sampling temperature $1.0$, top-$p$ $0.95$, and a maximum generation length of 38,912 tokens \citep{zhao2026opsd}. We evaluate our distillation checkpoints at steps 25, 50, 75, and 100 and report the highest score for each benchmark. 

In the results, B-OPSD denotes the version with static briefs, and B-OPSD$^\star$ denotes the version with refreshed briefs.

\subsection{Main Results}

Table~\ref{tab:main-results} shows that both B-OPSD variants have higher average accuracy than Base, SFT, GRPO, and OPSD at every model size tested. Compared with OPSD, B-OPSD gains 1.0, 0.6, and 0.6 percentage points at 1.7B, 4B, and 8B. B-OPSD$^\star$ gains 2.0, 1.3, and 1.3 points, respectively, and has the highest average at each size. These results show that problem briefs can provide useful teacher context without reference solutions during training in the reported settings.

Across the three benchmarks, B-OPSD$^\star$ achieves the highest average accuracy at every model size. On AIME 2024, both B-OPSD variants exceed OPSD at all three sizes. B-OPSD$^\star$ reaches 57.5 versus 54.7 for OPSD at 1.7B, and 77.5 versus 75.6 at 4B. On AIME 2025, B-OPSD$^\star$ reaches 70.3 versus 67.5 for OPSD at 4B. On HMMT 2025, B-OPSD exceeds OPSD at all three sizes, and B-OPSD$^\star$ reaches 50.0 versus 46.7 for OPSD at 8B.

Refreshing briefs raises the reported average over static briefs by 1.0, 0.7, and 0.7 points at 1.7B, 4B, and 8B. At 8B, refreshed briefs also achieve higher scores on AIME 2025 and HMMT 2025. Figure~\ref{fig:checkpoint-1p7b} shows the Qwen3-1.7B results across checkpoints. In its Average panel, B-OPSD$^\star$ exceeds both B-OPSD and OPSD at three of the four evaluated checkpoints. Briefs from later student checkpoints may provide teacher context that better reflects the student's changing policy than briefs fixed before training. These results show that problem briefs can improve self-distillation without verified reference solutions in the reported settings.

\subsection{Problem Brief Generation Strategies}

In this section, we examine whether problem briefs expected to be more accurate improve B-OPSD. A larger model may generate more accurate briefs, and access to a reference solution may also help the generator identify relevant information. We therefore compare two variants for distilling Qwen3-1.7B. In B-OPSD(8B), Qwen3-8B generates the briefs from the training problems. In B-OPSD(ref), the briefs are generated using the reference solutions supplied to OPSD. Both variants use the resulting briefs as privileged teacher context to train Qwen3-1.7B. B-OPSD(ref) uses reference solutions during brief preparation, while the main B-OPSD variants do not. All runs use the same training and evaluation settings for the 1.7B model. We report results on AIME 2024 and HMMT 2025 in Table~\ref{tab:brief-sources}.

\begin{table}[t]
  \centering
  \caption{Avg@12 accuracy (\%) for Qwen3-1.7B with different brief generation strategies. B-OPSD uses static briefs generated by the initial 1.7B model, and B-OPSD$^\star$ uses refreshed briefs. B-OPSD(8B) uses briefs generated by Qwen3-8B, while B-OPSD(ref) uses briefs generated using the reference solutions supplied to OPSD. Shaded rows show our main variants. Bold marks the best result in each column.}
  \begin{tabular}{lcc}
    \toprule
    \textbf{Method} & \textbf{AIME 2024} & \textbf{HMMT 2025} \\
    \midrule
    OPSD & 54.7 & 25.0 \\
    B-OPSD(8B) & 56.4 & 27.8 \\
    B-OPSD(ref) & 56.4 & 27.5 \\
    \midrule
    \rowcolor{bopsdrowblue} B-OPSD & 56.4 & 27.2 \\
    \rowcolor{bopsdrowblue} B-OPSD$^\star$ & \textbf{57.5} & \textbf{28.6} \\
    \bottomrule
  \end{tabular}
  \label{tab:brief-sources}
\end{table}

Both alternative brief sources improve on OPSD. B-OPSD(8B) and B-OPSD(ref) each match B-OPSD at 56.4 on AIME 2024. On HMMT 2025, they exceed B-OPSD by 0.6 and 0.3 points, respectively. B-OPSD$^\star$, which refreshes briefs using a recent student checkpoint, achieves the highest scores on both benchmarks. This pattern suggests that briefs adapted as the student learns may be more useful than briefs from a larger model or reference solutions.

\subsection{Effect of Divergence Choice}

Following OPSD \citep{zhao2026opsd}, we compare forward KL $D_{\mathrm{KL}}(p_T\|p_S)$, reverse KL $D_{\mathrm{KL}}(p_S\|p_T)$, and Jensen--Shannon divergence (JSD) with $\beta=0.5$ for B-OPSD$^\star$, B-OPSD, and OPSD on Qwen3-1.7B. We evaluate all three methods on AIME 2024 and HMMT 2025 and report Avg@12 accuracy, as in Table~\ref{tab:main-results}.

\begin{table}[t]
  \centering
    \caption{Avg@12 accuracy (\%) for Qwen3-1.7B with different divergences. B-OPSD uses static briefs; B-OPSD$^\star$ uses refreshed briefs. Bold marks the highest score in each row.}
  \begin{tabular}{lccc}
    \toprule
    \textbf{Divergence} & \textbf{OPSD} & \textbf{B-OPSD} & \textbf{B-OPSD$^\star$} \\
    \midrule
    \multicolumn{4}{l}{\textbf{AIME 2024}} \\
    Forward KL & 54.7 & 56.4 & \cellcolor{bopsdrowblue}\textbf{57.5} \\
    Reverse KL & 51.1 & 52.7 & \cellcolor{bopsdrowblue}\textbf{54.2} \\
    JSD ($\beta=0.5$) & 50.0 & 50.0 & \cellcolor{bopsdrowblue}\textbf{52.9} \\
    \midrule
    \multicolumn{4}{l}{\textbf{HMMT 2025}} \\
    Forward KL & 25.0 & 27.2 & \cellcolor{bopsdrowblue}\textbf{28.6} \\
    Reverse KL & 24.7 & 25.0 & \cellcolor{bopsdrowblue}\textbf{26.4} \\
    JSD ($\beta=0.5$) & 25.0 & 25.3 & \cellcolor{bopsdrowblue}\textbf{25.6} \\
    \bottomrule
  \end{tabular}
  \label{tab:divergence-choice}
\end{table}

Forward KL achieves the highest score for both B-OPSD variants on AIME 2024 and HMMT 2025. For OPSD, forward KL scores highest on AIME 2024 and ties JSD on HMMT 2025. Under each of the three divergences, B-OPSD$^\star$ has the highest score among the three methods on both benchmarks and exceeds OPSD. These results support the use of forward KL in our main Qwen3-1.7B experiments. They also show that B-OPSD$^\star$ can outperform OPSD across the tested divergences without using reference solutions during training.

\subsection{Problem Brief Component Ablation}

We examine the four sections of the problem brief using Qwen3-1.7B. Each ablation omits one section: targets, relevant principles, suggested approach, or pitfalls and checks. We evaluate both B-OPSD with static briefs and B-OPSD with refreshed briefs on AIME 2024 and HMMT 2025. Table~\ref{tab:brief-components} compares each ablation with the corresponding full brief.

\begin{table}[t]
  \centering
  \caption{Avg@12 accuracy (\%) for Qwen3-1.7B when one problem brief section is omitted. B-OPSD uses static briefs, and B-OPSD$^\star$ uses refreshed briefs. The shaded row uses the full brief. Bold marks the highest score within each variant and benchmark.}
  \setlength{\tabcolsep}{5pt}
  \begin{tabular}{lcccc}
    \toprule
    & \multicolumn{2}{c}{\textbf{B-OPSD}} & \multicolumn{2}{c}{\textbf{B-OPSD$^\star$}} \\
    \cmidrule(lr){2-3}\cmidrule(l){4-5}
    \textbf{Omitted section} & \textbf{AIME 2024} & \textbf{HMMT 2025} & \textbf{AIME 2024} & \textbf{HMMT 2025} \\
    \midrule
    \rowcolor{bopsdrowblue} None (full brief) & 56.4 & 27.2 & \textbf{57.5} & \textbf{28.6} \\
    Targets & 57.2 & \textbf{28.6} & 56.1 & 27.8 \\
    Relevant principles & 55.8 & 26.7 & 56.7 & 26.7 \\
    Suggested approach & \textbf{57.5} & 28.3 & 57.0 & 26.9 \\
    Pitfalls and checks & 56.7 & 26.7 & 56.7 & 28.3 \\
    \bottomrule
  \end{tabular}
  \label{tab:brief-components}
\end{table}

Relevant principles are the most consistent contributor in this ablation. Removing them lowers accuracy on both benchmarks for both brief variants. With refreshed briefs, AIME 2024 falls from 57.5 to 56.7 and HMMT 2025 falls from 28.6 to 26.7. The complete refreshed brief also achieves the highest score on both benchmarks within its variant; removing any of its four sections lowers accuracy. These results identify relevant principles as useful teacher context and show the value of retaining the complete brief when it is refreshed from student checkpoints.

\section{Conclusion}

We introduced B-OPSD, which generates a problem brief from the question and gives it to a frozen teacher as additional context. The current student samples solution trajectories from the question alone. The teacher evaluates the same trajectories and provides token-level distributions that guide student updates. This process uses no reference solutions during training and requires no separate teacher model. Across three Qwen3 model sizes and three competition mathematics benchmarks, both B-OPSD with static briefs and B-OPSD with refreshed briefs achieved higher average accuracy than OPSD in the reported runs. Refreshed briefs achieved the highest average at every model size. These results show that problem briefs can match or exceed the performance of OPSD without verified reference solutions during training. We will further examine how problem briefs should be updated during training, including when to refresh them. We will also test B-OPSD on reasoning tasks beyond competition mathematics.

{
    \small 
    \bibliographystyle{plain}
    \bibliography{custom}
}


\appendix

\FloatBarrier
\onecolumn
\clearpage
\raggedbottom
\section{Additional Implementation Details}
\label{sec:additional-implementation}

Our LoRA adapters cover the query, key, value, and output projections in attention and the gate, up, and down projections in the feed-forward layers. We use AdamW and bfloat16 precision. In the original OPSD study, SFT trains for 100 updates with a maximum sequence length of 16,000 tokens. GRPO trains for 500 updates and samples eight responses per problem at temperature $1.2$, with a maximum length of 16,000 tokens \citep{zhao2026opsd}.

The brief generation prompt asks for at most 700 tokens, but we set no separate hard length limit for brief decoding. We retain generated briefs without format filtering.

\section{Generated Problem Brief Examples}
\label{sec:brief-examples}

We randomly sampled candidate problem briefs generated by Qwen3-1.7B and selected three for illustration. Their formatting is adapted for LaTeX.

\begin{briefexample}{Example 1}

\noindent\textbf{Problem.} $n$ and $c$ are positive numbers such that increasing any positive number by $n\%$ and then decreasing the result by $c\%$ returns the original number. Prove that in this case
\[
\frac{n^{2}}{c^{2}}=\frac{100+n}{100-c}.
\]

\noindent\textbf{Generated brief.}

\noindent\textbf{1. Target and structure:}
The problem requires proving that $\frac{n^2}{c^2} = \frac{100 + n}{100 - c}$ given that applying $n\%$ increase followed by $c\%$ decrease returns the original number. The key is to derive a relationship between $n$ and $c$ from the condition that the operations are identity functions for all $x > 0$.

\noindent\textbf{2. Relevant principles:}
\begin{itemize}[leftmargin=1.25em,itemsep=1pt,topsep=2pt,parsep=0pt]
\item \textbf{Algebraic manipulation:} The condition that the operations return the original number leads to the equation $(1 + \frac{n}{100})(1 - \frac{c}{100}) = 1$.
\item \textbf{Solving equations:} Expanding and simplifying the equation to isolate $n$ in terms of $c$ (e.g., $n = \frac{100c}{100 - c}$) is critical.
\item \textbf{Substitution:} Substituting $n$ into the target expression to verify equality.
\end{itemize}

\noindent\textbf{3. Suggested approach:}
\begin{itemize}[leftmargin=1.25em,itemsep=1pt,topsep=2pt,parsep=0pt]
\item Start with the identity condition: $(1 + \frac{n}{100})(1 - \frac{c}{100}) = 1$.
\item Expand and simplify to derive $100(n - c) = nc$.
\item Solve for $n$ in terms of $c$: $n = \frac{100c}{100 - c}$.
\item Substitute $n$ into $\frac{n^2}{c^2}$ and $\frac{100 + n}{100 - c}$ to show they are equal.
\end{itemize}

\noindent\textbf{4. Pitfalls and checks:}
\begin{itemize}[leftmargin=1.25em,itemsep=1pt,topsep=2pt,parsep=0pt]
\item \textbf{Mistakes in algebra:} Ensure correct expansion of $(1 + \frac{n}{100})(1 - \frac{c}{100})$.
\item \textbf{Domain constraints:} Verify $100 - c \neq 0$ to avoid division by zero.
\item \textbf{Verification:} Confirm that both sides of the equality simplify to the same expression after substitution.
\end{itemize}

The proof relies on algebraic manipulation and substitution, ensuring the derived relationship holds for all $x > 0$.
\end{briefexample}

\begin{briefexample}{Example 2}
\noindent\textbf{Problem.} For what value of $m$ does the equation $(x+4)(x+1) = m + 2x$ have exactly one real solution? Express your answer as a common fraction.

\noindent\textbf{Generated brief.}

\noindent\textbf{Problem Brief:}

\noindent\textbf{Target and structure:}
Determine the value of $m$ for which the equation $(x+4)(x+1) = m + 2x$ has exactly one real solution. The main challenge is to find the value of $m$ that makes the equation have a unique real root.

\noindent\textbf{Relevant principles:}
The equation is a quadratic in $x$, and the number of real solutions depends on the discriminant of the quadratic. The discriminant $D$ of a quadratic equation $ax^2 + bx + c = 0$ is given by $D = b^2 - 4ac$. If $D > 0$, there are two distinct real solutions; if $D = 0$, there is exactly one real solution (a repeated root); and if $D < 0$, there are no real solutions.

\noindent\textbf{Suggested approach:}
\begin{itemize}[leftmargin=1.25em,itemsep=1pt,topsep=2pt,parsep=0pt]
\item Expand and rearrange the equation to standard quadratic form.
\item Identify the coefficients $a$, $b$, and $c$ of the quadratic.
\item Compute the discriminant $D$ and set it equal to zero to find the value of $m$ that results in exactly one real solution.
\item Verify the result by substituting back into the original equation.
\end{itemize}

\noindent\textbf{Pitfalls and checks:}
\begin{itemize}[leftmargin=1.25em,itemsep=1pt,topsep=2pt,parsep=0pt]
\item Ensure the equation is correctly expanded and rearranged.
\item Confirm that the discriminant is correctly calculated.
\item Verify that the value of $m$ leads to a repeated root by substituting back into the original equation.
\item Check that the discriminant condition is applied correctly to the quadratic form.
\end{itemize}
\end{briefexample}

\begin{briefexample}{Example 3}

\noindent\textbf{Problem.} Let $g$ be a graph with $n$ vertices and $m$ edges that contains no 4-cycle. Show that:
\[
m \leqslant \frac{n}{4}(1+\sqrt{4 n-3}).
\]

\noindent\textbf{Generated brief.}

\noindent\textbf{1. Target and structure:}
The problem asks to prove that a 4-cycle-free graph with $n$ vertices and $m$ edges satisfies $m \leq \frac{n}{4}(1+\sqrt{4n-3})$. The key challenge is to derive this bound using extremal graph theory principles, leveraging inequalities and combinatorial counting.

\noindent\textbf{2. Relevant principles:}
\begin{itemize}[leftmargin=1.25em,itemsep=1pt,topsep=2pt,parsep=0pt]
\item \textbf{Cauchy-Schwarz inequality:} Relates the sum of squares of degrees to the total number of edges.
\item \textbf{Counting paths of length 2:} In a 4-cycle-free graph, the number of paths of length 2 between any two vertices is at most 1, leading to a bound on the sum of degrees squared.
\item \textbf{Quadratic inequality derivation:} Combines the above to derive the upper bound on $m$.
\end{itemize}

\noindent\textbf{3. Suggested approach:}
\begin{itemize}[leftmargin=1.25em,itemsep=1pt,topsep=2pt,parsep=0pt]
\item Use the Cauchy-Schwarz inequality to relate $\sum d(v)^2$ to $m$:
\[
\sum d(v)^2 \geq \frac{(2m)^2}{n}.
\]
\item Count paths of length 2:
\[
\sum_{v} \binom{d(v)}{2} \leq \binom{n}{2}.
\]
\item Combine these to derive the quadratic inequality:
\[
4m^2 \leq 2m n + n^2(n-1),
\]
leading to the bound $m \leq \frac{n}{4}(1+\sqrt{4n-3})$.
\end{itemize}

\noindent\textbf{4. Pitfalls and checks:}
\begin{itemize}[leftmargin=1.25em,itemsep=1pt,topsep=2pt,parsep=0pt]
\item Ensure the counting of paths of length 2 is valid for all 4-cycle-free graphs.
\item Verify that the derived quadratic inequality correctly bounds $m$ without overcounting.
\item Confirm the asymptotic behavior of the bound matches known extremal graph theory results.
\end{itemize}

This approach connects combinatorial counting with inequality techniques to derive the required bound.
\end{briefexample}



\end{document}